\documentclass[10pt,letterpaper]{article}

\usepackage{ccn}
\usepackage{pslatex}
\usepackage{apacite}
\usepackage{amsfonts}
\usepackage[utf8]{inputenc}
\usepackage{amsfonts}
\usepackage{tikz}
\usepackage{svg}
\usepackage{comment}
\usetikzlibrary{arrows.meta}

\title{Optimizing deep video representation to match brain activity}
        
\author{{\large \bf Hugo Richard (hugo.richard@inria.fr)} \\
	PARIETAL Team, INRIA, 1 Rue Honor d'Estienne d'Orves, 91120 Palaiseau, France
  \AND {\large \bf Ana Luísa Pinho (ana-luisa.grilo-pinho@inria.fr)} \\
  PARIETAL Team, INRIA, 1 Rue Honor d'Estienne d'Orves, 91120 Palaiseau, France\\
  \AND {\large \bf Bertrand Thirion (bertrand.thirion@inria.fr)} \\
  PARIETAL Team, INRIA, 1 Rue Honor d'Estienne d'Orves, 91120 Palaiseau, France
  \AND {\large \bf Guillaume Charpiat (guillaume.charpiat@inria.fr)} \\
  TAU team, INRIA, LRI, Paris-Sud University, France
  \\}

\begin{document}

\maketitle

\section{Abstract}
{ 
The comparison of observed brain activity with the statistics
generated by artificial intelligence systems is useful to probe brain
functional organization under ecological conditions.
Here we study fMRI activity in ten subjects watching color natural
movies and compute deep representations of these movies with an architecture
that relies on optical flow and image content.
The association of activity in visual areas with the different layers
of the deep architecture displays complexity-related contrasts across
visual areas and reveals a striking foveal/peripheral dichotomy.
}
\begin{quote}
\small
\textbf{Keywords:}
deep learning; video encoding; brain mapping;
\end{quote}

\section{Introduction}
The understanding of brain functional architecture has long been
driven by subtractive reasoning approaches, in which the activation
patterns associated with different experimental conditions presented
in event-related or block designs are contrasted in order to yield
condition-specific maps \cite{poline2012}.
A more ecological way of stimulating subjects consists in presenting
complex continuous stimuli that are much more similar to every-day
cognitive experiences.

The analysis of the ensuing complex stimulation streams proceeds by
extracting relevant features from the stimuli and correlating the
occurrence of these features with brain activity recorded
simultaneously with the presentation of the stimuli.
The analysis of video streams has been carried in \cite{eickenberg2017seeing} 
or \cite{gucclu2015deep} using a deep convolutional network trained for 
image classification. 
More recently, \cite{gucclu2017increasingly} has used a deep neural network
trained for action recognition to analyze video streams.

Like \cite{gucclu2017increasingly}, we use a deep neural network trained for 
action recognition to extract video features and train a linear model to predict 
brain activity from these features. In contrast, our study is not restricted to dorsal
stream visual areas but involves the whole brain, and the deep neural network we use
is pretrained on the largest action recognition 
dataset available \cite{kay2017kinetics}.

From the different layers of the deep neural networks, we build video representations that allow us to segregate (1) occipital and lateral areas of the visual cortex (reproducing the results of \cite{gucclu2015deep})  and (2) foveal and peripheric areas of the visual cortex.  
We also introduce an efficient spatial compression scheme for deep video features that allows us to speed up the training of our predictive algorithm. We show that our compression scheme outperforms PCA by a large margin. 

\section{Methods}
\subsection{Deep video representation}
We use a deep neural network trained for action recognition to build deep representations of the  Berkeley
Video stimuli \cite{nishimoto2011reconstructing}.
This material consists of more than four hours of color natural movies
built by mixing video blocks of $5$-$15$ seconds in a random fashion.

The deep network we use is called Temporal Segment Network (TSN)
\cite{wang2016temporal}.
Following an idea introduced in 2014 \cite{simonyan2014two} it was intended to mimic
the dorsal and ventral stream by separately processing raw frames and
optical flow fields.
We chose TSN for our experiments because it uses a much larger number
of layers than the original network (which results in higher accuracy
in action recognition) and that a version of TSN pretrained on
Kinetics -- a massive video dataset (300~000 unique video clips) with
400 different classes all describing a human and at least 400 videos
per class -- is publicly available. The network is trained to recognize
human actions such as slack-lining, skateboarding, massaging feet, dancing zumba and dining.

The version of TSN we use in our experiments is based on Inception v3
\cite{szegedy2016rethinking} for both streams where small networks are
used as building blocks of the main large network
\cite{lin2013network}. 
Each stream in the TSN Network is composed of more than 40 convolution
layers and a fully connected layer.
The activities after the last layer represent the probability of
belonging to each action class.

\subsection{Feature extraction}
The raw frames encode information about pixels, and flow fields encode
information about pixels displacements.
Although flow fields and raw frames streams do not precisely
disentangle spatial content and motion information in videos, we may
expect that the raw frames stream better represent local spatial
features while the flow fields stream more efficiently convey dynamic
information.
Following \cite{eickenberg2017seeing} we consider that the activation
statistics in the first layers (the ones closer to those of the input) have
a low level of abstraction, whereas the last layers (closer to the
labels) represent high-level information. 
Therefore each activity in both streams can be considered as specific
features or representations of the video.

If we were to extract all network activities of the Berkeley Video
Dataset we would need to store more than 6 millions floats per frame
in the dataset.
Such a representation would be highly redundant. 
In order to keep the volume of data reasonable, in each stream we only focus on four convolutional layers $L_1, L_2, L_3, L_4$ ranked by complexity. 
We further compress the data using spatial smoothing, and use temporal
smoothing so that we get one representation every two seconds of
video, which allows us to match the acquisition rate of fMRI scanners.

\subsection{Regression}
10 subjects were scanned while watching the color natural movies of
the Berkeley Video Dataset.
The fMRI images were acquired at high spatial resolution (1.5mm), from
a Prisma Scanner, using Multi-band and IPAT accelerations (mb
factor=3, ipat=2).
These data are part of a large-scale mapping project on a limited
number of participants, called Human Brain Charting.
Data acquisition procedures and initial experiments run in this
project are described in \cite{ibc}.
In order to link extracted deep video features to the internal
representation of videos in each subject we use a simple
linear model to fit their brain activity in each voxel.

The use of a very simple model allows us to posit that the performance
of the predictive model from a particular video representation is
mostly linked to the suitability of the video representation.
Hence the performance of the algorithm can be seen as a measure of the
biological suitability of the video representation.

We use a kernel ridge regression with an hyper-parameter setting the
magnitude of the l2-penalization on the weights.
The resulting prediction is obtained using a cross validation
procedure (11 sessions are used for train, 1 for test and at least 5
different splits are considered).
To set the value of the hyper-parameter, we use a 5-fold cross
validation on the train set and consider 20 different values.
During hyper parameter selection, we only focus on the visual
cortex to make this computation efficient.

The chosen measure of performance of our prediction algorithm is the coefficient of determination $m_{cv}$. 
Let $\mathbf{y}_{pred}$ and $\mathbf{y}_{real}$ be the respectively the prediction of a voxel activity and the real voxel activity. Then
\[
m_{cv}(\mathbf{y}_{pred}, \mathbf{y}_{real}) \;=\; 1 - \frac{\sum_{t=1}^{n_b} (\mathbf{y}_{pred}[t] - \mathbf{y}_{real}[t])^2}{\sum_{t=1}^{n_b}(\mathbf{y}_{real}[t] - \overline{\mathbf{y}_{real}})^2}
 \] 
The metric used to select the best parameter is the number of voxels
having a coefficient of determination $m_{cv}$ greater than
$0.1$. This procedure leads to different parameter values depending on
the chosen layer activities.

\begin{figure}
\centering
\includegraphics[scale=0.4]{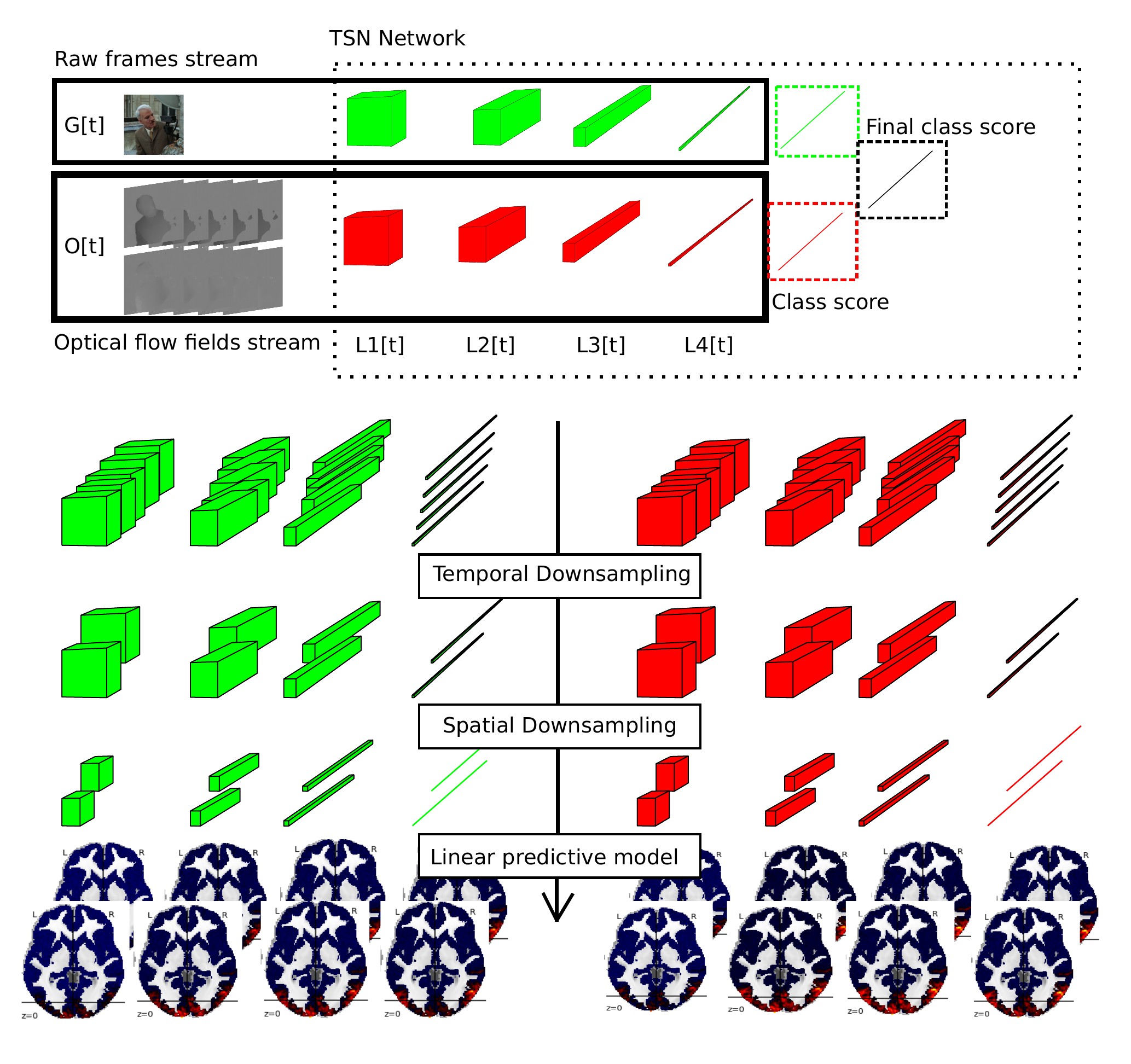}
\caption{ 
Feature extraction and regression scheme: at each time frame we
compute and extract the activities of four layers $L_1, \cdots, L_4$
of the temporal segment network on a single frame and on a stack of 5
consecutive optical flow fields.
The extracted activities are spatially and temporally down-sampled and
then used to predict brain activity of subjects exposed to the video
stimuli.}
\label{fig:feature_extraction}
\end{figure}

Figure~\ref{fig:feature_extraction} gives an overview of the pipeline
used to extract and process deep video features to estimate the brain
activity of subjects.

\section{Results}
The extracted deep network features lead to different prediction
performance depending on the down-sampling procedure, the stream used
and the localization of target voxels.

\subsection{An efficient spatial compression scheme}
We show that preserving the channel structure of the network during
spatial compression procedure is key for developing an efficient
compression scheme. 

We compare three spatial compression
schemes for network activities:
(1) Standard principal component analysis (PCA) with $2000$
components;
the transformation is learned on training sessions before it is
applied to all sessions.
(2) Average pooling inside channels (APIC) which computes local means
of activities located in the same channel.
(3) Average pooling inside and between convolution layers (APBIC)
which is used to get the same number of output features for all layers
while minimizing the number of convolutions between channels.
It allows us to check that the performance of the predictive algorithm
is not merely driven by the number of features.

The procedure for activities extraction, temporal down-sampling and
brain activity prediction is not changed while the spatial compression
scheme varies. 
The benchmark is performed using a leave-one-out cross-validation procedure with two splits in three subjects.

\begin{figure}
\hspace*{-3mm}
\includegraphics[scale=0.49]{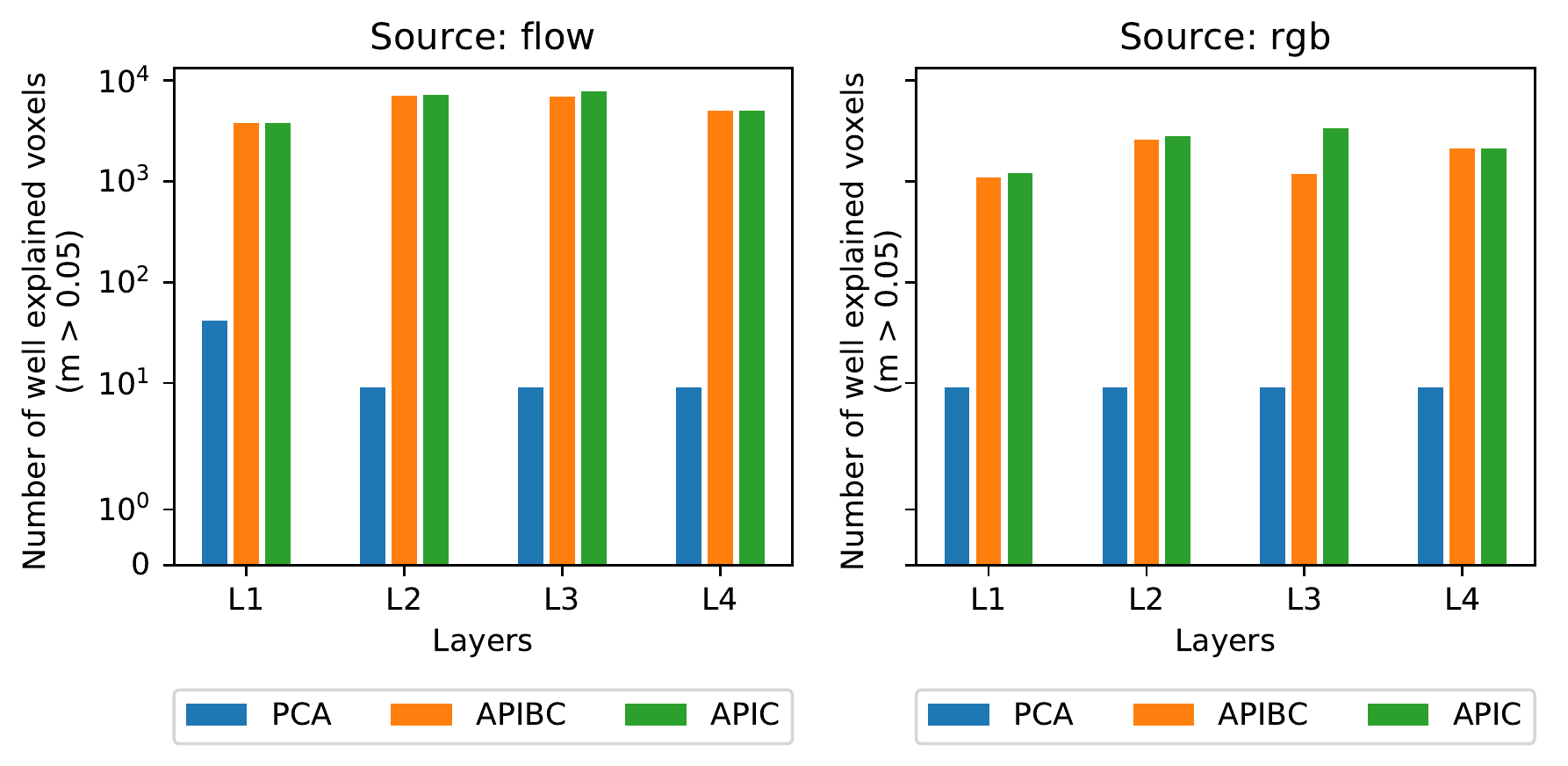}
\caption{ Comparison of the different neural network compression
  streams. The APIC approach slightly outperforms APIBC, and both
  strongly outperform PCA. When using APIC or APIBC we predict correctly up to 850 times more voxels than when using PCA.}
\label{fig:spatial_compression}
\end{figure}

Figure~\ref{fig:spatial_compression} shows that both approaches
preserving channel organization structure outperform PCA by a large
margin.

These results suggest that data stored in the same channel are similar
and that mixing data between channels tends to destroy valuable
information. In our pipeline, we average only inside same channels
(APIC) because it yields the best performance. 
Choosing APBIC would be trading performance for computation speed
since its high compression rate enables a much faster training of the
prediction algorithm.

\subsection{Data based parcellation of the brain using deep video representation}
Depending on the considered region of the brain, the best fitting
representation varies. We show that the compressed activities of different layers show
contrasts between low-level (retinotopic) versus high-level
(object-responsive) areas, but also between foveal and peripheral
areas.

\begin{figure}[t]
\centering
\includegraphics[scale=0.3]{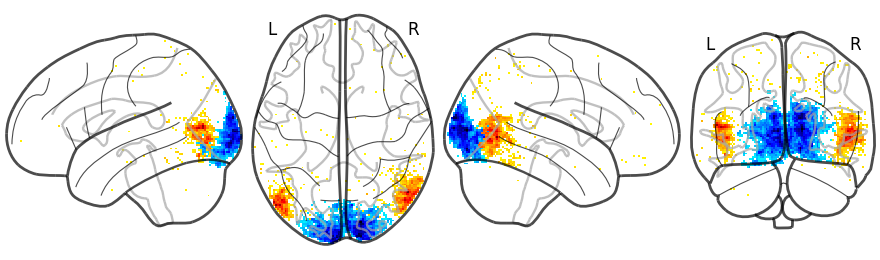}
\includegraphics[scale=0.3]{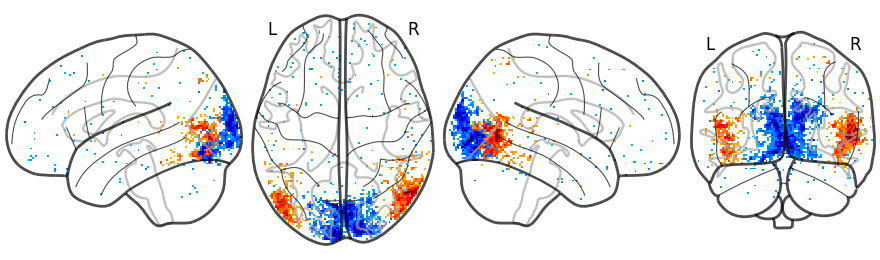}
\caption{
High level and low level areas contrasts: Difference between predictions score from high level layer activity and low level activity of the raw frames stream $L_4^{rgb} - L_2^{rgb}$ (top) and flow fields stream $L_4^{flow} - L_2^{flow}$ (bottom). 
The results show a clear contrast between occipital areas better
predicted by lower level layers (blue) and lateral areas better
predicted from highest level layers (red), illustrating a gradient of
complexity across areas.}
\label{fig:L4-L1}
\end{figure}

The difference between the prediction score from high level layer
activity and low level layer activity of both streams ($L_4^{flow} -
L_2^{flow}$ and $L_4^{rgb} - L_2^{rgb}$) yields a clear contrast
between occipital (low-level) and lateral (high-level) areas (see
Fig~\ref{fig:L4-L1}). This highlights a gradient of complexity in
neural representation along the ventral stream which was also found in
\cite{gucclu2015deep}.

\begin{figure}[t]
\centering
\begin{tabular}{cc}
\includegraphics[scale=0.30]{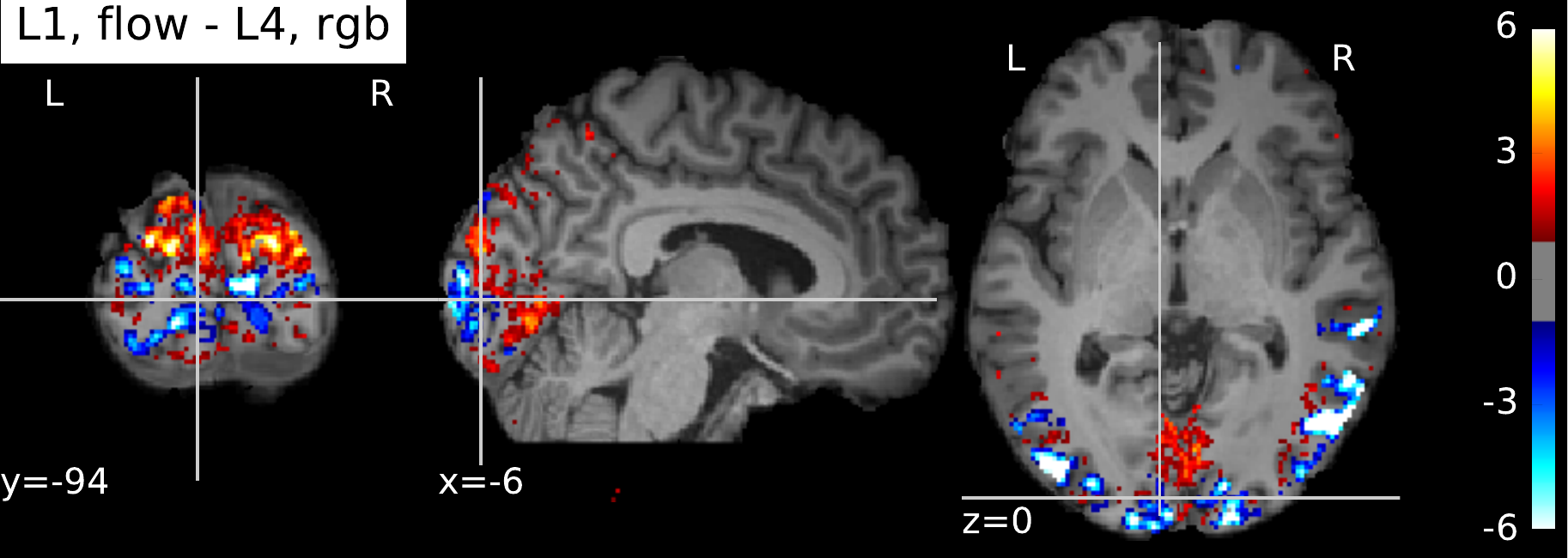}&\\
\includegraphics[scale=0.30]{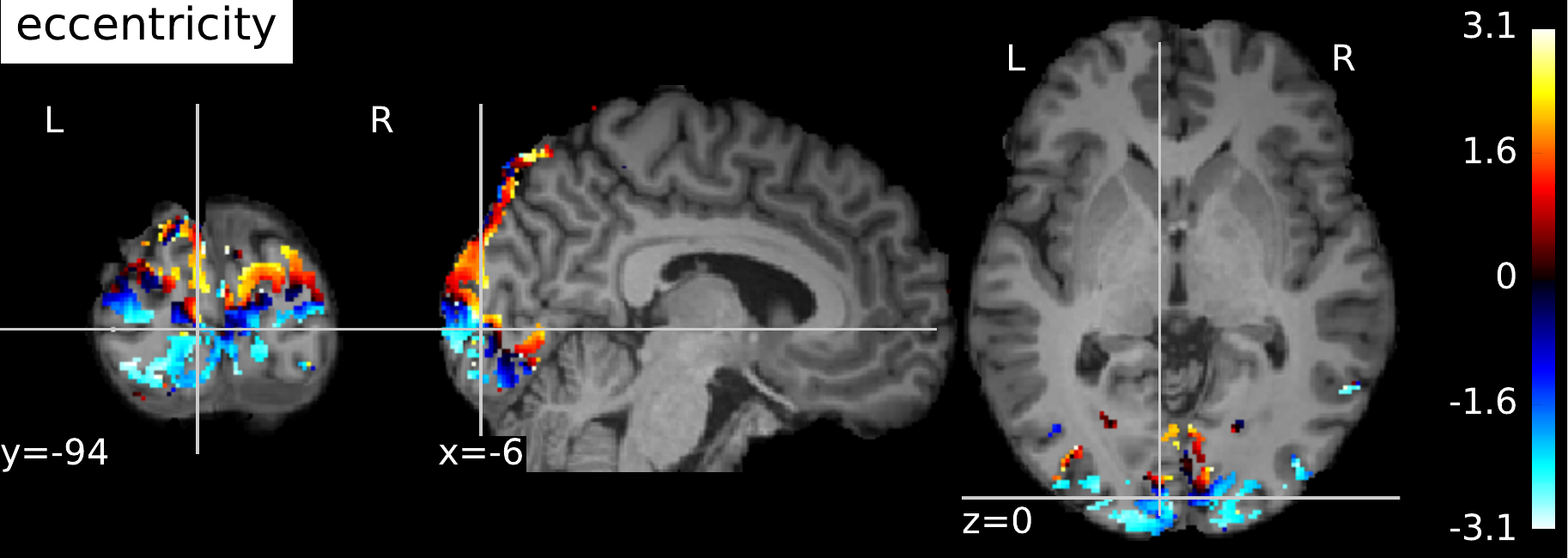}\\
\end{tabular}
\caption{
The difference between predictions score from low-level layers
activity of flow fields and high-level layers activity of raw frames
stream $L_1^{flow} - L_4^{rgb}$ (top) resembles the preferred
eccentricity map of the same subject (bottom).
Areas that are better predicted from low level flow fields streams
are mostly involved in peripheric vision whereas areas better predicted
from high level raw frames stream are mainly foveal.}
\label{fig:eccentricity-comparison}
\end{figure}

The difference between predictions score from low-level layers
activity of flow fields stream and high level layers activity of raw
frames stream ($L_1^{flow} - L_4^{rgb}$) yields a contrast that does
not match boundaries between visual areas; instead, it does coincide
with the retinotopic map displaying preferred eccentricity (see
Figure~\ref{fig:eccentricity-comparison}).
Intuitively this means that regions where brain activity is better
predicted from the highest layer of optical flow fields than from the
lowest layer of raw frames stream are involved in peripheric vision
whereas regions where activity is better predicted from the lowest layer
of raw frames stream than from the highest layer of optical flow
fields are mainly foveal.

\begin{figure}[t]
\centering
\includegraphics[scale=0.39]{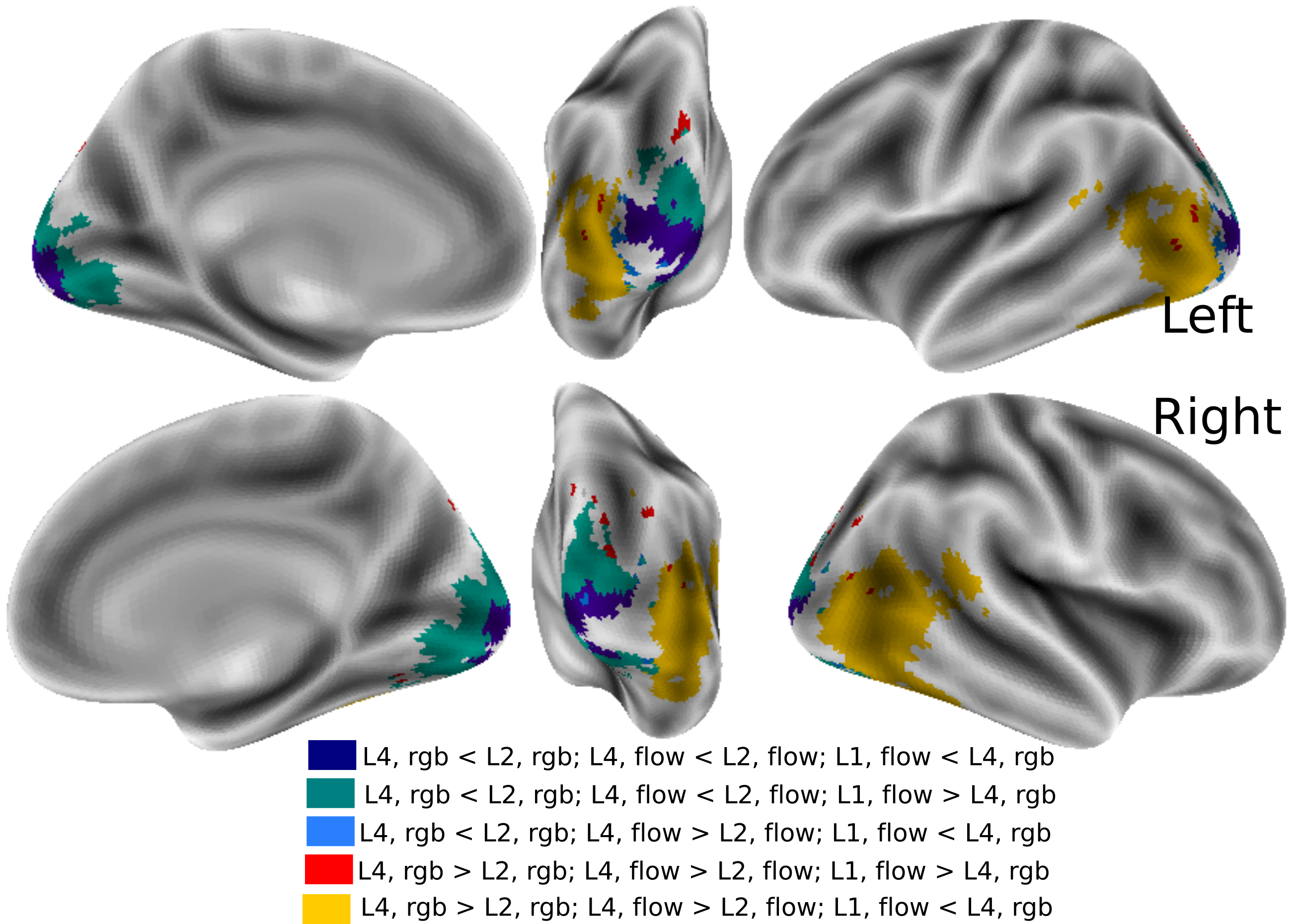}
\caption{Parcellation summarizing artificial-biological correspondences: 
The set of active voxels were split into subgroups according to their differential response to three contrasts:
$L2^{flow} - L4^{flow}$, $L2^{rgb} - L4^{rgb}$, and $L1^{flow}-L4^{rgb}$. 
From the 8 possible resulting profiles, 3 major clusters stand out:
deep blue, $L2^{flow} > L4^{flow}$, $L2^{rgb} > L4^{rgb}$, and
$L1^{flow} < L4^{rgb}$; it corresponds to a voxel set in primary
visual areas that has low eccentricity (foveal regions);
green, $L2^{flow} > L4^{flow}$, $L2^{rgb} > L4^{rgb}$, and $L1^{flow}
> L4^{rgb}$, it corresponds to the same visual areas, but for voxels
with higher eccentricity (peripheric voxels);
yellow , $L2^{flow} < L4^{flow}$, $L2^{rgb} < L4^{rgb}$, and $L1^{flow} <
L4^{rgb}$, it corresponds to lateral and lateral visual areas that
encode more abstract representations of the objects.  }
\label{fig:resulting-parcelisation.png}
\end{figure}

We use the contrasts between high level layers and low level layers,
and the eccentricity related contrast to construct a parcellation of
the brain based on these contrasts (see
Figure~\ref{fig:resulting-parcelisation.png}).
From the 8 possible resulting profiles, three major clusters stand out
allowing us to successfully depict a clustering of the voxels using
contrasts from deep representation of the stimuli.

\section{Discussion}
Reproducing the results of \cite{gucclu2015deep} we have shown that lateral areas are best predicted by the last layers of both streams whereas occipital areas are best predicted by first layers of both streams. We have also shown that foveal areas are best predicted by last layers of the raw frames stream and that peripheric areas are best predicted by the first layers of the flow fields stream.
We have introduced a compression procedure for video representation that does not alter too much the channel structure of the network, yielding tremendous gains in performance compared to PCA.

The linear prediction from deep video features yields predictions scores that are far better than chance. However the TVL1 algorithm \cite{zach2007duality} used in the TSN network does not produce high quality flow fields. Using more recent algorithms to compute optical flow such as Flownet 2 \cite{ilg2017flownet}, our performance could be further improved. The TSN Network would have to be retrained though.

In contrast to \cite{gucclu2017increasingly}, the data used to train
the network are not the same as the data presented to the subjects.
We rely in fact on transfer between computer vision datasets and the
visual content used for visual stimulation.
This transfer is imperfect: the Berkeley video dataset contains videos
of landscapes and animated pictures that are not present in the
Kinetic dataset, which introduces some noise.

In conclusion, our study provides key insights that areas have a role
linked to their retinotopic representation when performing action
recognition.
Future studies should focus on finessing this result by
using a network tuned for other tasks.

\section{Acknowledgments}
This project has received funding from the European Union’s Horizon
2020 Research and Innovation Programme under Grant Agreement
No. 720270 (HBP SGA1).

\bibliographystyle{apacite}

\setlength{\bibleftmargin}{.125in}
\setlength{\bibindent}{-\bibleftmargin}
\bibliography{biblio.bib}

\end{document}